%% file: main.tex
\documentclass[10pt,twocolumn,letterpaper]{article}

\usepackage[pagenumbers]{iccv} 


\definecolor{iccvblue}{rgb}{0.21,0.49,0.74}
\usepackage[pagebackref,breaklinks,colorlinks,allcolors=iccvblue]{hyperref}

\usepackage{graphicx}
\usepackage{booktabs}
\usepackage[misc]{ifsym}
\usepackage{lipsum}  
\usepackage[accsupp]{axessibility}  

\usepackage{listings}
\usepackage[ruled,linesnumbered]{algorithm2e}
\SetKwComment{Comment}{/* }{ */}
\usepackage{amsmath}  
\usepackage{multicol}
\usepackage{multirow}
\usepackage{makecell}
\usepackage{adjustbox}
\usepackage{bbding}
\usepackage{appendix}
\usepackage{orcidlink}
\usepackage{amsfonts}       
\usepackage{nicefrac}       
\usepackage{microtype}      
\usepackage{xcolor}         
\usepackage {pifont}
\usepackage{subcaption}
\usepackage{colortbl}
\definecolor{lightgray}{gray}{0.95}
\definecolor{color3}{gray}{0.95}
\definecolor{rouse}{rgb}{0.981,0.961,0.941}
\usepackage{float}
\usepackage{wrapfig}

\newcommand{\boldblue}[1]{\textcolor{blue}{\underline{#1}}}
\newcommand{\boldred}[1]{\textcolor{Red}{\textbf{#1}}}
\newcolumntype{T}{@{\hspace{0em}}c@{\hspace{0em}}}

\def\paperID{***} 
\def\confName{ICCV}
\def\confYear{2025}

\title{CTSR: Controllable Fidelity-Realness Trade-off Distillation for Real-World Image Super Resolution}

\author{
    Runyi Li$^{1}$\quad Bin Chen$^{1}$\quad Jian Zhang$^{1\dagger}$\quad Radu Timofte$^{2\dagger}$\\
    $^{1}$School of Electronic and Computer Engineering, Peking University, China \\
    $^{2}$Computer Vision Lab, CAIDAS \& IFI, University of W\"urzburg, Germany
}

\begin{document}
\maketitle
\renewcommand*{\thefootnote}{$\dagger$}
\footnotetext[1]{Corresponding author.}

\begin{abstract}
Real-world image super-resolution is a critical image processing task, where two key evaluation criteria are the fidelity to the original image and the visual realness of the generated results. Although existing methods based on diffusion models excel in visual realness by leveraging strong priors, they often struggle to achieve an effective balance between fidelity and realness.
In our preliminary experiments, we observe that a linear combination of multiple models outperforms individual models, motivating us to harness the strengths of different models for a more effective trade-off. Based on this insight, we propose a distillation-based approach that leverages the geometric decomposition of both fidelity and realness, alongside the performance advantages of multiple teacher models, to strike a more balanced trade-off.
Furthermore, we explore the controllability of this trade-off, enabling a flexible and adjustable super-resolution process, which we call CTSR (Controllable Trade-off Super-Resolution).
Experiments conducted on several real-world image super-resolution benchmarks demonstrate that our method surpasses existing state-of-the-art approaches, achieving superior performance across both fidelity and realness metrics. 
\end{abstract}
\begin{figure}
    \centering
    \includegraphics[width=1\linewidth]{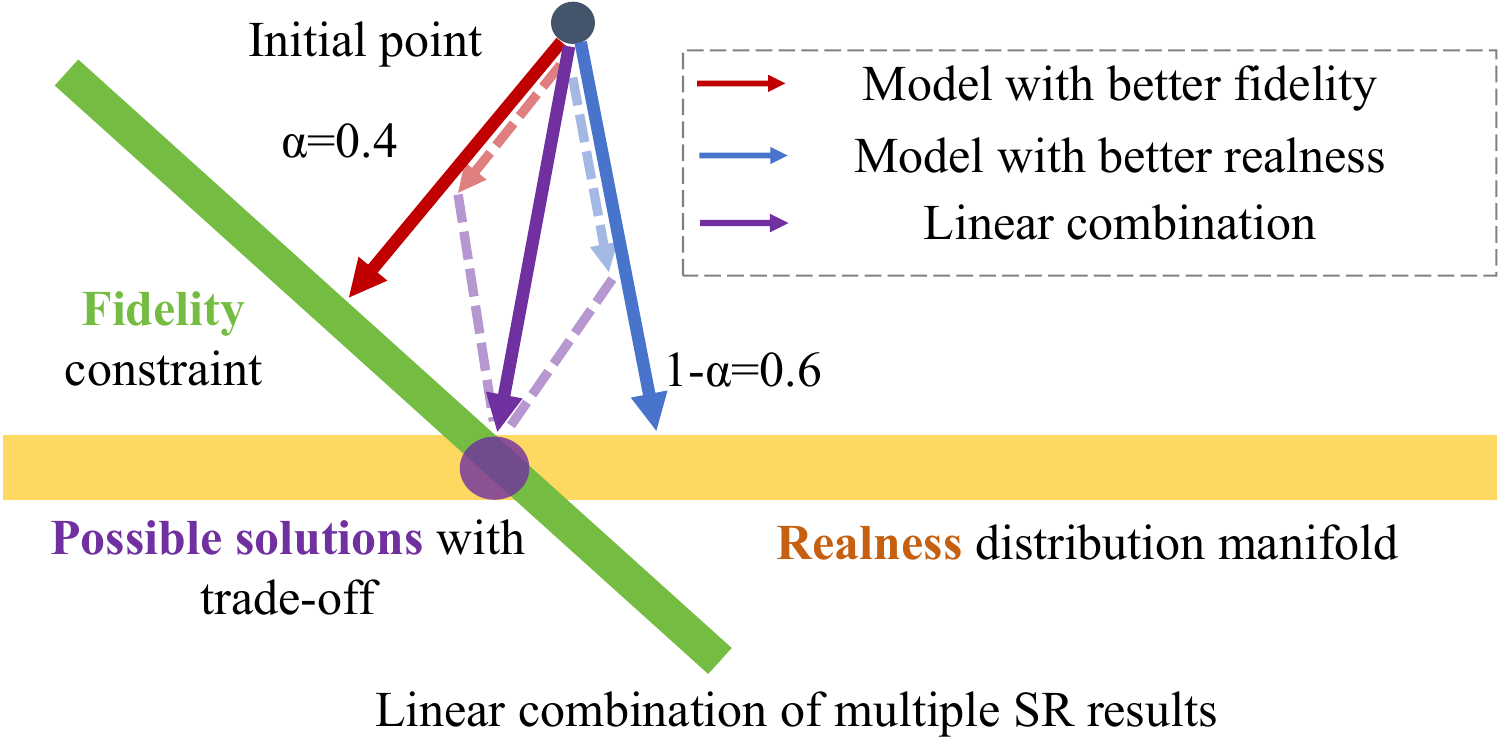}
    \caption{Illustration for vector decomposition in the image super-resolution process. It shows the simple linear approach, which serves as the \textbf{motivation} of our proposed CTSR.}
    \vspace{-3mm}
    \label{fig:motivation}
\end{figure}
\section{Introduction}
Image restoration, particularly image super-resolution (SR), is both a critical and challenging task in image processing. Early research~\cite{yang2010image,kim2010single,wang2015deep} typically focused on fixed degradation operators, such as downsampling and blur kernels, modeled as $\mathbf{y}=\mathbf{Ax}+\mathbf{n}$, where $\mathbf{x}$ represents the original image, $\mathbf{A}$ is the fixed degradation operator, $\mathbf{n}$ is random noise, and $\mathbf{y}$ is the degraded result. As the field has advanced, more recent work has shifted its focus to real-world degradation scenarios, where $\mathbf{A}$ turns to a complex and random combination of various degradations, with unknown degradation types and parameters.
The evaluation of image super-resolution is mainly based on two metrics: fidelity, which measures the consistency between the super-resolved image and the degraded image, and realness, which assesses how well the super-resolved image conforms to the prior distribution of natural images, as well as its visual quality~\cite{mentzer2020high,zhou2022quality,zhang2022spqe}.

The early methods primarily used architectures based on GAN~\cite{gan} and MSE, trained on pairs of original and degraded images~\cite{dong2015image, liang2021swinir, wang2018esrgan, guo2022lar}. These approaches excelled in achieving good fidelity in super-resolved results but often suffered from over-smoothing and detail loss~\cite{chen2024ccir}. The introduction of diffusion models brought powerful visual priors to the SR task, significantly improving the realness and visual quality of super-resolved images. However, these models frequently struggle with maintaining consistency between the super-resolved and degraded images. Achieving a satisfactory balance between fidelity and realness remains a challenge, with most methods failing to strike an effective trade-off.

Our insight is inspired by the simple attempt at linear combination. Let the super-resolved result with good fidelity be $\mathbf{x}_f$, and the super-resolved result with good realness be $\mathbf{x}_r$. By linearly combining them as 
$\mathbf{x}_c = \alpha \mathbf{x}_f + (1-\alpha) \mathbf{x}_r$, we can manually adjust $\alpha \in [0,1]$ to achieve a better balance between fidelity and realness in the final result $\mathbf{x}_c$.
Building on the insights from DDS~\cite{dds}, we treat the super-resolution process based on diffusion as a vector in the manifold space~\cite{chung2022improving,soh2019natural}, from the low-resolution (LR) input to the high-resolution (HR) output. This vector can be geometrically decomposed into two components: (1) \textit{convergence} toward the natural image distribution using the diffusion prior, ensuring \textbf{realness}, and (2) \textit{correction} toward consistency constraints, ensuring \textbf{fidelity}. This decomposition is illustrated in Fig.~\ref{fig:motivation}.

Motivated by this observation, we propose a \textbf{controllable} trade-off real-world image super-resolution method based on fidelity-realness distillation, which we name CTSR. Our method distills a diffusion-based SR approach with high fidelity to an existing SR model with strong realness, which also serves as a teacher to distill itself, maintaining its superior performance of realness.
Furthermore, To achieve a continuous and controllable trade-off, we further distill the model using the flow-matching technique~\cite{lipman2022flow,zhu2024oftsr,fischer2023boosting}, enabling it to freely adjust between fidelity and realness. Specifically, assuming sampling steps range from $0$ to $T$, the distilled model exhibits better fidelity at step $T$, better realness at step $0$, and a trade-off between the two at intermediate steps. Based on this approach, our CTSR enables the controllability of SR results, as shown in Fig.~\ref{fig:teaser}. To summarize, our contributions are three-fold:
    
\noindent \ding{113}~We propose a real-world image super-resolution method based on fidelity-realness distillation, effectively achieving a trade-off between fidelity and realness.

\noindent \ding{113}~We further introduce a continuous and controllable trade-off approach through another distillation process, enabling the model to freely adjust the balance between fidelity and realness, thus providing practical user flexibility and advancing the optimization of image SR tasks.

\noindent \ding{113}~Experiments on real-world image SR benchmarks demonstrate the superior performance of our proposed CTSR method, along with efficient inference sampling steps and reduced trainable parameter count.

\section{Related Work}
\noindent \textbf{GAN-based and MSE-oriented Image SR Methods}
Earlier work mainly use GAN~\cite{gan} and MSE-oriented~\cite{Vaswani_Shazeer_Parmar_Uszkoreit_Jones_Gomez_Kaiser_Polosukhin_2017, dong2015image} networks to implement the image SR task~\cite{ren2020real, wang2021realesrgan, pan2021exploiting, wang2018esrgan, wang2023gpsr, poirier2023robust}. 
SRGAN~\cite{ledig2017photo} first uses the GAN network to image SR task, optimized via both GAN and perceptual losses, to improve visual quality. Based on this observation, ESRGAN~\cite{wang2018esrgan} improved detail recovery by incorporating a relativistic average discriminator. Methods like BSRGAN~\cite{Zhang_2021_ICCV} and Real-ESRGAN~\cite{wang2021realesrgan} follow the complexities of real-world degradation, allowing the ISR approaches to effectively tackle uncertain degradation, thus improving the flexibility of the model. Although GAN-based methods can inject more realistic detail into images, they struggle with challenges such as training instability.
For MSE-oriented methods, SwinIR~\cite{liang2021swinir} introduces a strong baseline model for image restorations, which includes image super-resolution (including known degradation and real-world types), image denoising, and JPEG compression artifacts. As this method is also trained in an end-to-end manner, it also faces problems like over-smooth and detail missing.

\noindent \textbf{Diffusion-based Image SR Methods}
As diffusion models have developed, their strong visual priors have also been applied to image super-resolution tasks. SR3~\cite{sr3} first proposes a diffusion model for SR task, which uses LR input as the condition of diffusion sampling, thus needs training for the UNet. Further methods like DDRM~\cite{kawar2022denoising}, DDNM~\cite{wang2022zero} and DPS~\cite{chung2023diffusion} use classifier-free guidance~\cite{ho2022classifier}, which takes LR input as the guidance of original diffusion sampling; thus, these methods are training-free. However, all these methods are on a fixed degradation setting, where the degradation type and parameters are known.

As these training-free methods use gradient guidance to correct the diffusion sampling process, methods such as DiffBIR~\cite{diffbir} and GDP~\cite{fei2023generative} try to leverage the gradient to update the parameters of the degradation operator, and in this case the degradation type is known but the parameters are unknown.
Current diffusion-based image SR methods mainly focus on the real-world scenario, where the degradation is unknown and complex~\cite{StableSR_Wang_Yue_Zhou_Chan_Loy_2023, xie2024addsr, wu2024seesr, wang2024sinsr, wu2024osediff, yue2024resshift, yang2023pasd, yu2024scaling}. StableSR~\cite{StableSR_Wang_Yue_Zhou_Chan_Loy_2023} proposes an image SR method based on Stable Diffusion~\cite{Rombach_Blattmann_Lorenz_Esser_Ommer_2022}, using an adapter to introduce the LR guidance for diffusion sampling. However, such approach needs multiple steps to obtain SR result, which is time-consuming. ResShift~\cite{yue2024resshift} designs a special sampling, accelerating the overall sampling in 15 steps. Currently some methods try to distill the diffusion-based SR methods into one step, including AddSR~\cite{xie2024addsr}, SinSR~\cite{wang2024sinsr} and OSEDiff~\cite{yu2024scaling}. There are also some papers explore the controllability of diffusion-based SR, including PiSA-SR~\cite{sun2024pixel} and OFTSR~\cite{zhu2024oftsr}.

\section{Preliminaries}
\noindent \textbf{Diffuion Probablistic Models}
\cite{ho2020denoising,song2021denoising,song2020score} are a class of generative models with strong visual prior. The key idea is to model the data distribution by simulating a forward noise-adding process and a reverse denoising process.
Let $\mathbf{x}_{0}$ represent the original image, $\mathbf{x}_{t}$ be the data at the t-th step of the forward process. The forward process can be described as:
$q(\mathbf{x}_{t}|\mathbf{x}_{t-1})=\mathcal{N}(\mathbf{x}_{t} ; \sqrt{1-\beta_{t}} \mathbf{x}_{t-1}, \beta_{t} \mathbf{I})$, where $\beta_t$ controls the noise added at each step, and  $\mathcal{N}(\cdot,\boldsymbol{\mu},\sigma^2\mathbf{I})$ represents Gaussian distribution with mean $\mu$ and co-variance matrix $\sigma^2\mathbf{I}$.
The reverse process aims to reconstruct the original data $\mathbf{x}_0$ by predicting $\mathbf{x}_{t-1}$ from $\mathbf{x}_t$:
$p_{\theta}(\mathbf{x}_{t-1} | \mathbf{x}_{t})=\mathcal{N}(\mathbf{x}_{t-1} ; \boldsymbol{\mu}_{\theta}(\mathbf{x}_{t}, t), \sigma_{t}^{2} \mathbf{I})$, where $\boldsymbol{\mu}_\theta(\mathbf{x}_t,t)$ is the predicted mean parameterized by a neural network. 

The training of the diffusion model needs a reconstruction loss of the difference between added noise in forward process, and predicted noise in reverse process, formulated as 
$L= \sum_{t=1}^{T}[||\boldsymbol{\epsilon}_\theta(\mathbf{x}_t,t)-\boldsymbol{\epsilon}||^2]$, where $\boldsymbol{\epsilon}_\theta(\mathbf{x}_t,t)$ is the model's prediction of the noise $\epsilon$ added at each timestep.
\begin{figure*}
    \centering
    \includegraphics[width=0.99\linewidth]{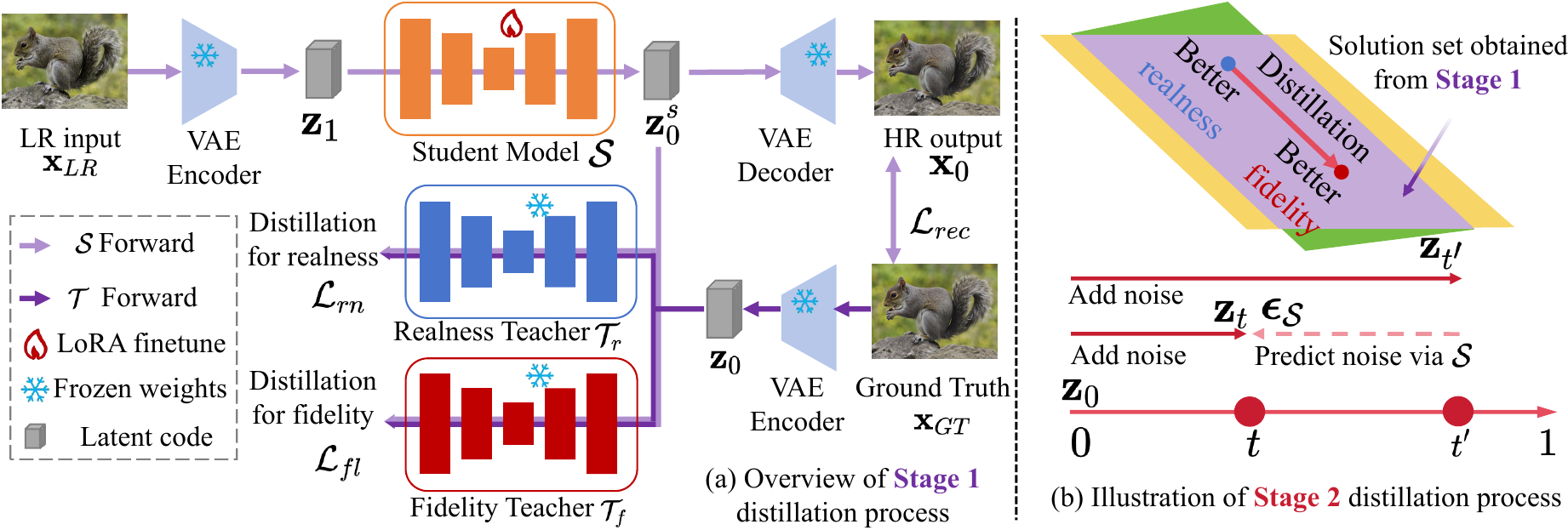}
    \caption{Illustration of our proposed CTSR. (a) At the first stage, we distill student model via two teacher models, one with better fidelity performance, and one with better realness performance. (b) At the second stage, we distill model obtailed from first stage, to a continuous mapping to SR results with different trade-offs between fidelity and realness.}
    \label{fig:ovewview}
    \vspace{-2mm}
\end{figure*}

Flow matching~\cite{liu2022flow,lipman2022flow} is a generative modeling technique similar to diffusion models~\cite{meng2023distillation}. It can model and learn the mapping from one data distribution to another through a noise-adding and denoising process, similar to diffusion models. Such distribution transformation process can be applied to tasks such as image reconstruction and style transfer~\cite{martin2024pnp,dao2023flow,hu2024latent,yin2024one}.

\noindent \textbf{Convex Optimization for Image Restoration}
Image restoration, when modeled as $\mathbf{y}=\mathbf{Ax}+\mathbf{n}$, is also known as image inverse problem. The target for image restoration is as $\mathop{\arg\min}\limits_\mathbf{x}||\mathbf{y}-\mathbf{Ax}||_2^2+\lambda \mathcal{R}(\mathbf{x})$, where $\mathcal{R}(\mathbf{x})$ is the regularization term, like $L_1$ norm or total variation~\cite{rudin1992nonlinear,zhu2024perceptual}.
This convex optimization problem can be solved via algorithms like gradient descent and ISTA~\cite{ito2019trainable}, in an iterative process. Take gradient descent step as an example:
$ \mathbf{x}_{k+1}=\mathbf{x}_k + \rho \nabla_{\mathbf{x}}(\mathbf{y}-\mathbf{Ax}_k)$, where $\mathbf{x}_{k}$ and $\mathbf{x}_{k+1}$ is the restoration result in $k$ and $k+1$ step, and $\rho$ is the learning rate. Diffision-based image SR methods, like DPS~\cite{chung2023diffusion} and DDS~\cite{dds}, are inspired via such process, taking iterative sampling in diffusion as optimization steps.

\section{Method}
\subsection{Motivation}
\label{sec:motivation}
In diffusion-based methods, some approaches excel in fidelity, such as ResShift~\cite{yue2024resshift} and SinSR~\cite{wang2024sinsr}, while others prioritize realness metrics, like OSEDiff~\cite{wu2024osediff} and StableSR~\cite{StableSR_Wang_Yue_Zhou_Chan_Loy_2023}. Combining the strengths of these methods can facilitate an effective trade-off between the two. One straightforward approach is to linearly combine the super-resolved outputs of different models. For example, by multiplying the image tensor of ResShift by $\alpha$ and OSEDiff by $(1-\alpha)$, and then summing them, both fidelity and realness metrics can be improved by adjusting the coefficients. We validate this on the Nikon test subset of RealSR~\cite{cai2019toward}, with the results shown in Tab.~\ref{tab:linear}. We further interpret this linear combination method as the sum of vectors corresponding to different SR methods in image space, as illustrated in Fig.~\ref{fig:motivation}.

\begin{table}
    \centering
    \begin{tabular}{ccccc}
    \toprule
        Settings & PSNR$\uparrow$ & LPIPS$\downarrow$ & Inference time (s) \\
        \hline
        $\alpha=0$   &  24.54 & 0.3575 & 0.7546\\
        $\alpha=0.2$ &  24.84 & \boldblue{0.3525} & 0.9196\\
        $\alpha=0.4$ &  25.25 & 0.3633 & 0.9196\\
        $\alpha=0.6$ &  \boldblue{25.34} & 0.3742 & 0.9196\\
        $\alpha=0.8$ &  25.10 & 0.3857 & 0.9196\\
        $\alpha=1.0$   &  24.88 & 0.3915 & \boldred{0.1791}\\
        \hline
        Ours   &  \boldred{25.45} & \boldred{0.3411} & \boldred{0.1791}\\
    \bottomrule
    \end{tabular}
    \caption{Results of the linear combination on RealSR~\cite{cai2019toward} Nikon sub-testset. $\alpha$ is multiplied with ResShift~\cite{yue2024resshift}, and $(1-\alpha)$ with OSEDiff~\cite{wu2024osediff}. By adding SR results from two models, the performance for both fidelity and realness is improved. Best and second-best results shown in \boldred{red} and \boldblue{blue}.}
    \vspace{-5mm}
    \label{tab:linear}
\end{table}
However, the performance of the linear combination method above is limited, and its inference speed is slower due to the need to run two models. To address these issues and enhance the model's representation capability, we extend it to a more general framework.
Inspired by the success of knowledge distillation in image SR~\cite{hui2019lightweight,zhang2021data,zhang2024pairwise,zhu2024information}, we distill the model output to the intersection of consistency constraints and high-quality image distribution manifolds, striking a trade-off of fidelity and realness.
To further enable controllability of the trade-off between fidelity and realness, we distill the diffusion sampling process of the model into a transformation from realness to fidelity, allowing for a flexible, controllable adjustment between the two. 
As a result, users can freely adjust these two properties according to their preferences in practical scenarios.
\subsection{Overview}
\label{sec:overview}
Our model is an one-step diffusion-based SR approach finetuned from OSEDiff~\cite{wu2024osediff}. The training scheme consists of two stages.
In the first stage, as shown in Fig.~\ref{fig:ovewview}(a), we select an SR model with good realness as the student model $\mathcal{S}$. This model is distilled via LoRA~\cite{hu2022lora} using two teacher models: one with high fidelity (denoted as $\mathcal{T}_f$) and another with good realness (denoted as $\mathcal{T}_r$). The teacher model $\mathcal{T}_f$ guides the student model $\mathcal{S}$ with gradient directions for fidelity, while $\mathcal{T}_r$ ensures that the student model retains its original generative capability. As a result, the super-resolution process of the model receives gradient corrections in the fidelity direction, and converges to the intersection of the fidelity constraint and the realness distribution manifold.

In the second stage, as shown in Fig.~\ref{fig:ovewview}(b), we further distill $\mathcal{S}$ within the solution set obtained from the first stage. Since the diffusion model can be viewed as a distribution transformation mapping from the initial input to the final output, we set the starting point as the super-resolved result from the first stage, with the target transformation being the solution with better fidelity within the solution set. This distribution transformation is achieved through distillation. 
As the time step $t$ of the diffusion model is continuous, we can controllably select the appropriate trade-off state, allowing us to achieve better and more diverse super-resolution results. An illustration of our proposed CTSR is shown in Fig.~\ref{fig:ovewview}.
\subsection{Stage 1: Distillation via Dual-Teacher Learning}
\label{sec:stage1}
Motivated by the insight in Sec.~\ref{sec:motivation}, we propose a distillation-based method, where two super-resolution models with good fidelity, $\mathcal{T}_f$, and realness, $\mathcal{T}_r$, are used to distill the original model $\mathcal{S}$. Our training objective consists of two components:

\noindent \textbf{Reconstruction Loss}. The output of the student model should be consistent with the original model in terms of both consistency and visual quality. We choose $L_2$ loss and LPIPS loss as the reconstruction loss terms: 
\begin{equation}
    \mathcal{L}_{rec} = \lambda_{l2} ||\mathcal{S}(\mathbf{x}_{LR})-\mathbf{x}_{GT}||_2^2 + \lambda_{lp} \ell (\mathcal{S}(\mathbf{x}_{LR}), \mathbf{x}_{GT})
\end{equation}
, where $\mathbf{x}_{LR}$ is input LR image, $\mathbf{x}_{GT}$ is ground-truth image, $\ell$ is LPIPS loss, $\lambda_{l2}$ and $\lambda_{lp}$ are balancing hyper-parameters.

\noindent \textbf{Dual Teacher Distillation Loss}. For ease of implementation, we use the same model for both the realness teacher $\mathcal{T}_r$ and the student model $\mathcal{S}$. This allows us to  split the distillation process into two parts:
(1) The fidelity teacher model $\mathcal{T}_f$ guides the gradients of $\mathcal{S}$, adjusting its output distribution toward a more faithful direction.
(2) The realness teacher model $\mathcal{T}_r$ regulates the student model, ensuring that the directional correction in (1) does not deviate from the manifold of the true image distribution achieved by $\mathcal{T}_r$. 
The specific formula for $\mathcal{L}_{fl}$ is as follows: 
\begin{equation}
\begin{aligned}
    \mathcal{L}_{fl}
    & =
    ||\boldsymbol{\epsilon}_{\mathcal{T}_{f}}\left (\mathbf{z}_{t}^s,t,c\right ) -
    \boldsymbol{\epsilon}_{\mathcal{S}}\left (\mathbf{z}_{t}^s,t,c\right )||_2^2 \\
    & + \gamma_{time}
    ||\boldsymbol{\epsilon}_{\mathcal{T}_{f}}\left (\mathbf{z}^{s}_{t},t,c\right ) -
    \boldsymbol{\epsilon}_{\mathcal{T}_{f}}\left (\mathbf{z}_{t},t,c\right )||_2^2,
\end{aligned}
\end{equation}
where $\boldsymbol{\epsilon}_{\mathcal{T}_{f}}$ and $\boldsymbol{\epsilon}_{\mathcal{S}}$ represent the denoising UNet of $\mathcal{T}_{f}$ and $\mathcal{S}$, respectively; $c$ is prompt embedding;
$\mathbf{z}_{t}$ and $\mathbf{z}^{s}_{t}$ are the latent codes of ground-truth $\mathbf{x}_{GT}$ and the student model’s SR result $\mathbf{x}_{0}$, obtained via VAE encoder $\mathcal{E}$, each added with the noise at timestep $t$ in the forward process of the diffusion model; 
$\gamma_{time}$ is the hyperparameter for balancing the two terms. 
The first term $\boldsymbol{\epsilon}_{\mathcal{T}_{f}}\left (\mathbf{z}_{t}^s,t,c\right ) -
    \boldsymbol{\epsilon}_{\mathcal{S}}\left (\mathbf{z}_{t}^s,t,c\right )$ aligns the output of $\mathcal{S}$ with the teacher model $\mathcal{T}_{f}$, enabling the student model to learn the distribution information from the teacher. 
The second term, $\boldsymbol{\epsilon}_{\mathcal{T}_{f}}\left (\mathbf{z}^{s}_{t},t,c\right ) - \boldsymbol{\epsilon}_{\mathcal{T}_{f}}\left (\mathbf{z}_{t},t,c\right )$, leverages the teacher model's prior to align the SR result $\mathbf{x}_{0}$ with $\mathbf{x}_{GT}$. 
Since the alignment in the second term is achieved by adding noise to the latent codes of $\mathbf{x}_{0}$ and $\mathbf{x}_{GT}$ separately, and calculating the difference in the predicted noise of $\mathcal{T}_{f}$, it reflects the distributional difference between them in the image space. As a result, compared to directly using $L_2$ loss, this approach better captures the distributional differences between the student model and the ground truth, avoiding issues like over-smoothing and loss of detail typically introduced by $L_2$ loss, while preserving the semantic details of the original image. We depict the detailed calculating process of $\mathcal{L}_{fl}$ in \textbf{\textcolor{blue}{Supplementary Materials}}.

This design is similarly applied for the distillation of $\mathcal{T}_r$:
\begin{equation}
\begin{aligned}
    \mathcal{L}_{rn}
    & =
    ||\boldsymbol{\epsilon}_{\mathcal{T}_{r}}\left (\mathbf{z}_{t}^s,t,c\right ) -
    \boldsymbol{\epsilon}_{\mathcal{S}}\left (\mathbf{z}_{t}^s,t,c\right )||_2^2 \\
    & + \gamma_{time}
    ||\boldsymbol{\epsilon}_{\mathcal{T}_{r}}\left (\mathbf{z}^{s}_{t},t,c\right ) -
    \boldsymbol{\epsilon}_{\mathcal{T}_{r}}\left (\mathbf{z}_{t},t,c\right )||_2^2,
\end{aligned}
\end{equation}


By combining these losses, the student model $\mathcal{S}$ can achieve improved fidelity without sacrificing its original performance. As a result, the linear combination method discussed in Sec.~\ref{sec:motivation} is extended to a more general approach, where the student's convergence direction evolves from a simple vector sum to a more precise optimal solution direction. 
This distillation mechanism is inspired by the SDS~\cite{poole2022dreamfusion} and VSD~\cite{wang2023prolificdreamer,dong2024tsd} losses, which regulate the student model using both the teacher model and the ground truth. 

The loss function for distillation in the first stage is:
\begin{equation}
    \mathcal{L}_{s1} = \mathcal{L}_{rec} + \lambda_{rn} \mathcal{L}_{rn} + \lambda_{fl} \mathcal{L}_{fl},
\end{equation}
where $\lambda_{rn}$ and $\lambda_{rn}$ are balancing weights. 

In short, our proposed distillation method guides student model $\mathcal{S}$ toward the intersection of the fidelity constraint and the realness distribution. The distilled SR model then serves as the teacher model in the following second stage, providing SR solutions with fidelity-realness trade-off.

\begin{table*}[!t]

\resizebox{0.99\textwidth}{!}{%
\begin{tabular}{@{}c|c|ccccccccc@{}}
\toprule
\textbf{Datasets} & \textbf{Method} & \textbf{PSNR $\uparrow$} & \textbf{SSIM $\uparrow$} & \textbf{LPIPS $\downarrow$ } & \textbf{DISTS $\downarrow$} & \textbf{FID $\downarrow$} & \textbf{NIQE $\downarrow$} & \textbf{MUSIQ $\uparrow$} & \textbf{MANIQA $\uparrow$} & \textbf{CLIPIQA $\uparrow$} \\ \hline

 & RealESRGAN~\cite{wang2021realesrgan} & \boldblue{28.62} & \boldblue{0.8052} & 0.5428 & \boldblue{0.2374} & \boldblue{171.79} & 7.8675 & 54.26 & \boldblue{0.5202} &  0.4515 \\
 & ResShift~\cite{yue2024resshift} & \boldred{28.69} & 0.7874 & \boldred{0.3525} & 0.2541 & 176.77 & 7.8762 & 52.40 & 0.4756 & 0.5413 \\
 & SinSR~\cite{wang2024sinsr} & 28.38 & 0.7497 & 0.3669 & 0.2484 & 172.72 & \boldred{6.9606} & \boldblue{55.03} & 0.4904 & \boldblue{0.6412} \\
\multirow{-4}{*}{\textbf{DRealSR}} & CTSR (\it{t}=0.8) (ours) & 28.47 & \boldred{0.8056} & \boldblue{0.3561} & \boldred{0.2369} & \boldred{161.24} & \boldblue{7.8462}  & \boldred{58.76} & \boldred{0.5453} & \boldred{0.6745} \\ \hline

 & RealESRGAN~\cite{wang2021realesrgan} & 25.69 & \boldred{0.7614} & 0.3266 & \boldblue{0.1646} & 168.02 & \boldred{4.0146} & 60.36 & 0.3934 & 0.4495 \\
 & ResShift~\cite{yue2024resshift} & \boldred{26.39} & \boldblue{0.7567} & \boldred{0.3158} & 0.2432 & 149.59 & 6.8746 & 60.22 & \boldblue{0.5419} & 0.5496 \\
 & SinSR~\cite{wang2024sinsr} & 26.27 & 0.7351 & 0.3217 & 0.2341 & \boldblue{137.59} & 6.2964 & \boldblue{60.76} & 0.5418 & \boldblue{0.6163} \\
\multirow{-4}{*}{\textbf{RealSR}} & CTSR (\it{t}=0.2) (ours) & \boldblue{26.29} & 0.7211 & \boldblue{0.3210} & \boldred{0.1620} & \boldred{127.67} & \boldblue{4.2979} & \boldred{66.84} & \boldred{0.6314} & \boldred{0.6435} \\ \hline

 & RealESRGAN~\cite{wang2021realesrgan} & 24.29 & 
 \boldred{0.6372} & 0.3570 & \boldblue{0.1621} & 46.31 & \boldred{3.4591} & 61.05 & 0.3830 & 0.5276 \\
 & ResShift~\cite{yue2024resshift} & \boldred{24.71} & \boldblue{0.6234} & 0.3473 & 0.2253 & 42.01 & 6.3615 & 60.63 & 0.5283 & 0.5962 \\
 & SinSR~\cite{wang2024sinsr} & 24.41 & 0.6018 & \boldred{0.3262} & 0.2068 & \boldblue{35.55} & 5.9981 & \boldblue{62.95} & \boldblue{0.5430} & \boldblue{0.6501} \\
\multirow{-4}{*}{\textbf{DIV2K-Val}} & CTSR (\it{t}=0.2) (ours) & \boldblue{24.45} & 0.6098 & \boldblue{0.3384} & \boldred{0.1394} & \boldred{24.75} & \boldblue{3.6803} & \boldred{69.25} & \boldred{0.5826} & \boldred{0.6726} \\ \bottomrule
\end{tabular}%
}
\caption{
Quantitative comparison with the state-of-the-art (SOTA) methods, which have superior performance on \textit{fidelity}. ``Ours-t" here is chosen as the results when timestep is set as $t$. The best and second-best results of each metric are highlighted in \boldred{red} and \boldblue{blue}.}
\label{tab:fidelity}
\vspace{-1mm}
\end{table*}

\begin{table*}[!t]
\resizebox{0.99\textwidth}{!}{%
\begin{tabular}{@{}c|c|ccccccccc@{}}
\toprule
\textbf{Datasets} & \textbf{Method} & \textbf{PSNR $\uparrow$} & \textbf{SSIM $\uparrow$} & \textbf{LPIPS $\downarrow$ } & \textbf{DISTS $\downarrow$} & \textbf{FID $\downarrow$} & \textbf{NIQE $\downarrow$} & \textbf{MUSIQ $\uparrow$} & \textbf{MANIQA $\uparrow$} & \textbf{CLIPIQA $\uparrow$} \\ \hline

  & StableSR~\cite{StableSR_Wang_Yue_Zhou_Chan_Loy_2023} & \boldred{28.04} & 0.7454 & \boldblue{0.3279} & 0.2272 & 144.15 & \boldblue{6.5999} & 58.53 & 0.5603 & 0.6250 \\
 & DiffBIR~\cite{diffbir} & 25.93 & 0.6525 & 0.4518 & 0.2761 & 177.04 & \boldred{6.2324} & \boldred{65.66} & \boldblue{0.6296} & 0.6860 \\
 & SUPIR~\cite{yu2024scaling} & 25.09 & 0.6460 & 0.4243 & 0.2795 & 169.48 & 7.3918 & 58.79 & 0.5471 & 0.6749 \\
 & PASD~\cite{yang2023pasd} & \boldblue{27.79} & 0.7495 & 0.3579 & 0.2524 & 171.03 & 6.7661 & 63.23 & 0.5919 & 0.6242 \\
 & InvSR~\cite{yue2024arbitrary} & 26.75 & 0.6870 & 0.4178 & \boldblue{0.2144} & 142.98 & 6.7030 & 63.92 & 0.5439 & 0.6791 \\
 & OSEDiff~\cite{wu2024osediff} & 27.35 & \boldblue{0.7610} & \boldred{0.3177} & 0.2365 & \boldred{141.93} & 7.3053 & 63.56 & 0.5763 & \boldblue{0.7053} \\
\multirow{-7}{*}{\textbf{DRealSR}} & CTSR (\it{t}=0.0) (ours) & 27.38 & \boldred{0.7767} & 0.3423 & \boldred{0.1937} & \boldblue{142.52}  & 6.6438 & \boldblue{64.70} & \boldred{0.6412} & \boldred{0.7060} \\ \hline

 & StableSR~\cite{StableSR_Wang_Yue_Zhou_Chan_Loy_2023} & 24.62 & 0.7041 & 0.3070 & 0.2156 & 128.54 & 5.7817 & 65.48 & 0.6223 & 0.6198 \\
 & DiffBIR~\cite{diffbir} & 24.24 & 0.6650 & 0.3469 & 0.2300 & 134.56 & 5.4932 & \boldred{68.35} & \boldblue{0.6544} & \boldblue{0.6961} \\
 & SUPIR~\cite{yu2024scaling} & 23.65 & 0.6620 & 0.3541 & 0.2488 & 130.38 & 6.1099 & 62.09 & 0.5780 & 0.6707 \\
 & PASD~\cite{yang2023pasd} & \boldblue{25.68} & \boldred{0.7273} & 0.3144 & 0.2304 & 134.18 & 5.7616 & \boldblue{68.33} & 0.6323 & 0.5783 \\
 & InvSR~\cite{yue2024arbitrary} & 24.50 & \boldblue{0.7262} & \boldred{0.2872} & \boldblue{0.1624} & 148.16 & \boldblue{4.2189} & 67.45 & \boldred{0.6636} & 0.6918 \\
 & OSEDiff~\cite{wu2024osediff} & 23.94 & 0.6736 & 0.3172 & 0.2363 & \boldblue{125.93} & 6.3822 & 67.52 & 0.6187 & \boldred{0.7001} \\
\multirow{-7}{*}{\textbf{RealSR}} & CTSR (\it{t}=0.0) (ours) & \boldred{25.70} & 0.6962 & \boldblue{0.3058} & \boldred{0.1530} & \boldred{121.30} & \boldred{4.0662} & 67.94 & 0.6367 & 0.6495 \\ \hline

 & StableSR~\cite{StableSR_Wang_Yue_Zhou_Chan_Loy_2023} & 23.27 & 0.5722 & \boldblue{0.3111} & 0.2046 & \boldblue{24.95} & 4.7737 & 65.78 & \boldblue{0.6164} & 0.6753 \\
 & DiffBIR~\cite{diffbir} & 23.13 & 0.5717 & 0.3469 & 0.2108 & 33.93 & 4.6056 & 68.54 & \boldred{0.6360} & 0.7125 \\
 & SUPIR~\cite{yu2024scaling} & 22.13 & 0.5279 & 0.3919 & 0.2312 & 31.40 & 5.6767 & 63.86 & 0.5903 & \boldblue{0.7146} \\
 & PASD~\cite{yang2023pasd} & \boldblue{24.00} & 0.6041 & 0.3779 & 0.2305 & 39.12 & 4.8587 & 67.36 & 0.6121 & 0.6327 \\
 & InvSR~\cite{yue2024arbitrary} & 23.32 & 0.5901 & 0.3657 & \boldred{0.1370} & 28.85 & \boldred{3.0567} & \boldblue{68.97} & 0.6122 & \boldred{0.7198} \\
 & OSEDiff~\cite{wu2024osediff} & 23.72 & \boldred{0.6109} & \boldred{0.3058} & 0.2138 & 26.34 & 5.3903 & 65.27 & 0.5838 & 0.6558 \\
\multirow{-7}{*}{\textbf{DIV2K-Val}} & CTSR (\it{t}=0.0) (ours) & \boldred{24.34} & \boldblue{0.6093} & 0.3377 & \boldblue{0.1377} & \boldred{24.56} & \boldblue{3.5455} & \boldred{69.52} & 0.5894 & 0.6741 \\ \bottomrule
\end{tabular}%
}
\caption{Quantitative comparison with SOTA methods having better performance on \textit{realness}. ``Ours-0.0" here denotes the result when timestep of CTSR is set as $0$. The best and second-best results of each metric are highlighted in \boldred{red} and \boldblue{blue}, respectively.}
\label{tab:realness}
\vspace{-2mm}
\end{table*}

\subsection{Stage 2: Distillation for Controllablility}
\label{sec:stage2}

As shown in Fig.~\ref{fig:motivation}, we represent the optimal solution as a point within a set of feasible solutions obtained from from Sec.~\ref{sec:stage1}. 
Within this set, some solutions exhibit better realness, while others demonstrate superior fidelity. Let the possible solutions be denoted as $\mathbf{x}_0$ and $\mathbf{x}_1$, where $\mathbf{x}_0$ performs better in realness and $\mathbf{x}_1$ excels in fidelity. These solutions can be viewed as distinct points in the high-dimensional space (where the image domain is considered as a $channel \times height \times width$ space), each at varying distances from the degradation constraint and the ground truth distribution manifold.

Building on the flow matching approach~\cite{lipman2022flow,zhu2024oftsr}, we further distill $\mathcal{S}$, mapping the diffusion sampling process from $\mathbf{x}_0$ to $\mathbf{x}_1$ within the feasible solution set. The input timesteps $t$ for the diffusion UNet in this mapping act as an adjustable parameter, enabling a a continuous, controllable trade-off between fidelity and realness.

Our training approach involves adding different noise to $\mathbf{z}_0$, the latent code of $\mathbf{x}_{0}$, and fine-tuning the denoising UNet $\boldsymbol{\epsilon}_{\mathcal{S}}$. The noise addition process, starting from $t=0$, is as follows:
\begin{equation}
    \mathbf{z}_{t} = \mathbf{z}_{0} + t \boldsymbol{\epsilon}_{\mathcal{T}_{S}}(\mathbf{z}_{0},t,c),
\end{equation}
where $\mathcal{T}_{S}$ denotes the student model obtained from the first stage.
This process also applies for timesteps starting at $t$:
\begin{equation}
    \mathbf{z}_{t'} = \mathbf{z}_{t} + (\Delta t ) \boldsymbol{\epsilon}_{\mathcal{T}_{S}}(\mathbf{z}_{0},t,c),
\end{equation}
where $\Delta t=t'-t$.
The student model $\mathcal{S}$ should adhere to the following equation:
\begin{equation}
\label{eq:regularization}
    t'\epsilon_{\mathcal{T}_S}(\mathbf{z}_0,t',c)
    -t\epsilon_{\mathcal{T}_S}(\mathbf{z}_0,t,c)=
    (\Delta t) \epsilon_{\mathcal{S}}(\mathbf{z}_t,t,c),
\end{equation}
wher the left-hand term represents the difference in added noise to latent code, and the right-hand term is the predicted noise by the student model $\mathcal{S}$. To satisfy the requirement in Eq.~\ref{eq:regularization}, $\mathcal{S}$ is finetuned using the following loss:
\begin{equation}
\begin{aligned}
    \mathcal{L}_{ctrl_{t,t'}} & = ||t\boldsymbol{\epsilon}_{\mathcal{T}_s}(\mathbf{z}_0,t,c) - t'\boldsymbol{\epsilon}_{\mathcal{T}_s}(\mathbf{z}_0,t',c) \\
    & + (\Delta t )
    \boldsymbol{\epsilon}_{\mathcal{S}}(\mathbf{z}_{t},t,c) ||_2^2,
\end{aligned}
\end{equation}
where $t$ and $t'$ are uniformly sampled from the range 0 to 1, separately. The overall loss function for controllable trade-off distillation in the second stage is then:
\begin{equation}
    \mathcal{L}_{s2} = \mathcal{L}_{ctrl_{t,t'}}, \; t,t' \in (0,1).
\end{equation}

By randomly sampling the timesteps $t$ and $t'$, the student model gradually learns the distribution of the first-stage solution set, acquiring information about solutions with better fidelity. Due to the inherent uncertainty and diversity in the noise addition and UNet predictions during distillation, our model efficiently utilizes the diversity of the diffusion model in SR tasks.



\begin{figure*}
    \centering
    \includegraphics[width=1.0\linewidth]{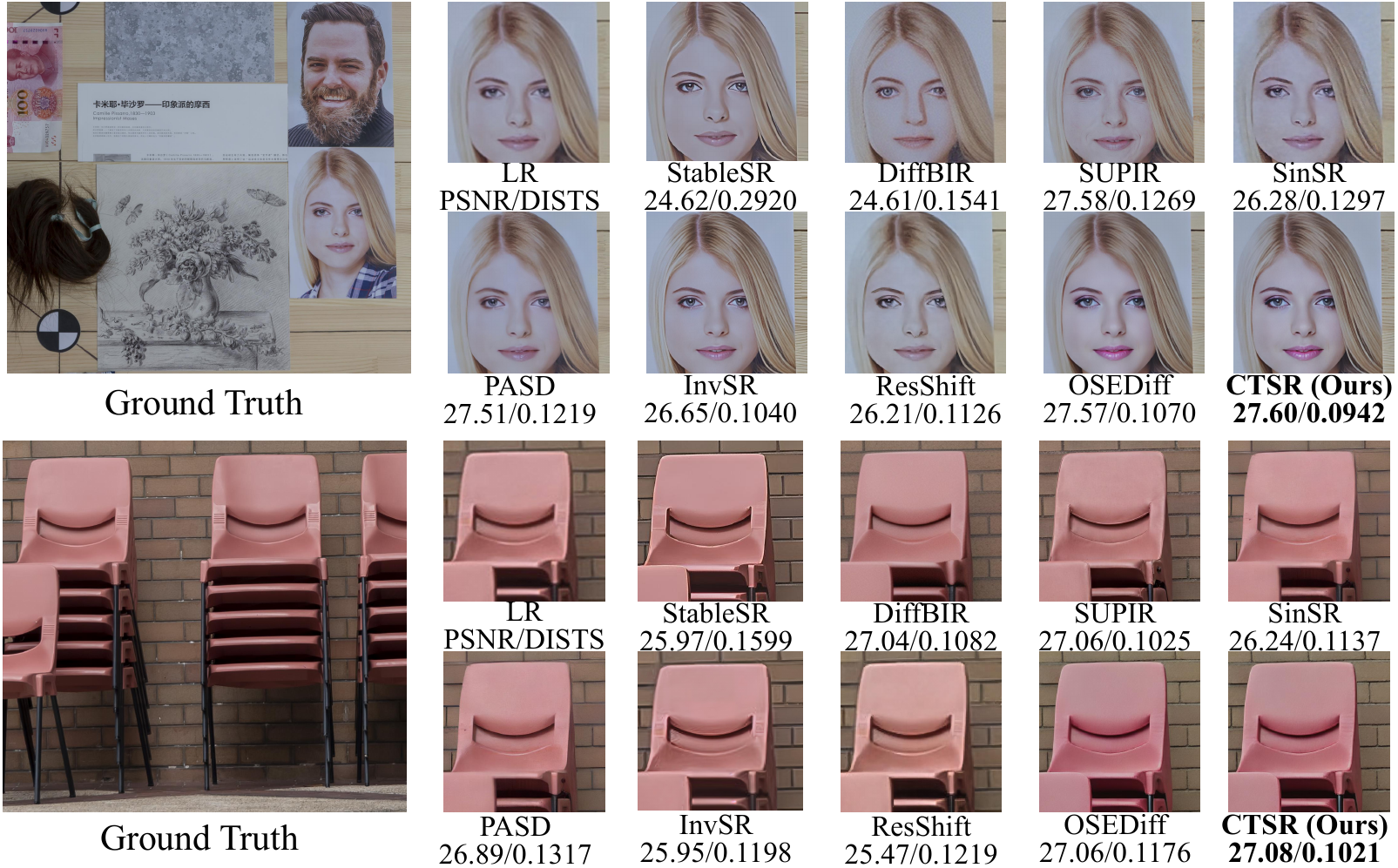}
    \caption{Visualized results of evaluation on the RealSR testset, with our proposed CTSR ($t=0.0$) and compared methods.}
    \label{fig:comp}
    \vspace{-4mm}
\end{figure*}


\section{Experiments}
\subsection{Settings}
\noindent \textbf{Datasets} We merge the training sets from 
DIV2K~\cite{agustsson2017ntire}, LSDIR~\cite{li2023lsdir}, DRealSR~\cite{wei2020component}, ImageNet~\cite{deng2009imagenet}, and RealSR~\cite{cai2019toward} as our training dataset, and evaluate our method on the validation sets of DIV2K, DRealSR, and RealSR. Degraded images are generated using the real-world degradation operator from RealESRGAN~\cite{wang2021realesrgan}.

\noindent \textbf{Evaluation Metrics} We assess both fidelity and realness for the super-resolution task. For fidelity, we use PSNR and SSIM~\cite{wang2004image}; for realness, we use LPIPS~\cite{zhang2018unreasonable}, DISTS~\cite{ding2020image}, and FID~\cite{heusel2017gans}, which require reference images, and NIQE~\cite{zhang2015feature}, MUSIQ~\cite{ke2021musiq}, CLIPIQA~\cite{wang2023exploring}, and MANIQA~\cite{yang2022maniqa}, which are reference-free. LPIPS uses VGG~\cite{simonyan2014very} weights following ~\cite{dong2024tsd}, and MANIQA uses PIPAL~\cite{jinjin2020pipal} weights by default.

\noindent \textbf{Implementation Details} For the teacher model selection, we choose OSEDiff~\cite{wu2024osediff} as $\mathcal{T}_{r}$, due to its advantage in realness, and ResShift~\cite{yue2024resshift} as$\mathcal{T}_{r}$, for its better fidelity performance. The pretrained version of Stable Diffusion~\cite{Rombach_Blattmann_Lorenz_Esser_Ommer_2022} used is 2.1-base. The default image input size for the models is 512$\times$512. All images are processed at their original size, and for images larger than 512$\times$512, we use patch splitting and apply VAE tiling to avoid block artifacts. In both the first and second stages of training, we use the AdamW~\cite{loshchilov2017decoupled} optimizer with $\beta_1$=0.9, $\beta_2$=0.999, and a learning rate of 5e-5, with 20,000 training steps in the first stage and 50,000 in the second stage. The batch size is set to 1. Distilltion in both stages is performed using LoRA~\cite{hu2022lora} fine-tuning, with a rank of 4.
For the loss balancing coefficients in $\mathcal{L}_{s1}$, $\lambda_{rn}$ is set to 1, $\lambda_{fl}$ to 2, and $\gamma_{time}$ to 5.5. In $\mathcal{L}_{rec}$, $\lambda_{l2}$ and $\lambda_{lp}$ are set to 1 and 2 respectively. 
All experiments are conducted on SR task with a scaling factor of 4, using an NVIDIA A6000 GPU.

\subsection{Comparison with State-of-the-arts}
\noindent \textbf{Comparison Methods}. We select methods for comparison based on two performance metrics: fidelity and realness, and group them accordingly. For fidelity, we choose ResShift~\cite{yue2024resshift}, SinSR~\cite{wang2024sinsr}, and RealESRGAN~\cite{wang2021realesrgan}; for realness, we select StableSR~\cite{StableSR_Wang_Yue_Zhou_Chan_Loy_2023}, DiffBIR~\cite{diffbir}, SUPIR~\cite{yu2024scaling}, SinSR~\cite{wang2024sinsr}, PASD~\cite{yang2023pasd}, InvSR~\cite{yue2024arbitrary}, and OSEDiff~\cite{wu2024osediff}.
\begin{table}
    \centering
    \resizebox{1\linewidth}{!}{
    \begin{tabular}{c|ccccc}
    \toprule
         \textbf{Loss type} & \textbf{PSNR}$\uparrow$ & \textbf{SSIM}$\uparrow$ & \textbf{LPIPS}$\downarrow$ & \textbf{CLIPIQA}$\uparrow$ & \textbf{MANIQA}$\uparrow$ \\
         \hline
         w/o distill & \boldred{26.71} & 0.6743 & 0.4552 & 0.5439 & 0.5775\\
         w/ distill (Ours) & 25.70 & \boldred{0.6962} & \boldred{0.3058} & \boldred{0.6495} & \boldred{0.6367}\\
         
    \bottomrule
    \end{tabular}
    }
    \caption{Ablation of training with and without dual teacher distillation loss. Best results are shown in \boldred{red}.}
    \label{tab:wot}
    \vspace{-2mm}
\end{table}

\begin{table}
    \centering
    \resizebox{0.95\linewidth}{!}{
    \begin{tabular}{c|ccccc}
    \toprule
        \textbf{Teacher} $\mathcal{T}_{fl}$ & \textbf{PSNR}$\uparrow$ & \textbf{SSIM}$\uparrow$ & \textbf{LPIPS}$\downarrow$ & \textbf{CLIPIQA}$\uparrow$ & \textbf{MANIQA}$\uparrow$ \\
    \hline
        SinSR & \boldred{25.71} & 0.6734 & 0.3552 & 0.6036 & 0.6065 \\
        ResShift (Ours) & 25.70 & \boldred{0.6962} & \boldred{0.3058} & \boldred{0.6495} & \boldred{0.6367}\\
    \bottomrule
    \end{tabular}
    }
    \caption{Results of different choices on $\mathcal{T}_{fl}$, evaluated on the RealSR testset. Best results are shown in \boldred{red}.}
    \label{tab:sinsr}
    \vspace{-3mm}
\end{table}

\begin{table}
    \centering
    \resizebox{0.99\linewidth}{!}{
    \begin{tabular}{c|cc| c|cc| c|cc}
    \toprule
        $\lambda_{rn}$ & \textbf{PSNR}$\uparrow$ & \textbf{LPIPS}$\downarrow$  & 
        $\lambda_{fl}$ & \textbf{PSNR}$\uparrow$ & \textbf{LPIPS}$\downarrow$  &
        $\gamma_{time}$ & \textbf{PSNR}$\uparrow$ & \textbf{LPIPS}$\downarrow$  \\
        \hline
        0.6 & 25.07 & 0.3487  & 1.6 & 25.81 & 0.3377  & 4.5 & 25.08 & 0.3481 \\
        0.8 & 24.81 & \boldblue{0.3185}  & 1.8 & \boldblue{25.62} & 0.3365  & 5.0 & 25.60 & \boldblue{0.3166} \\
        \textbf{1.0} & \boldred{25.70} & \boldred{0.3058}  & \textbf{2.0} & \boldred{25.70} & \boldred{0.3058}  & \textbf{5.5} & \boldblue{25.70} & \boldred{0.3058} \\
        1.2 & \boldblue{25.66} & 0.3376  & 2.2 & 25.44 & \boldblue{0.3149}  & 6.0 & 24.82 & 0.3212 \\
        1.4 & 25.62 & 0.3317  & 2.4 & 25.19 & 0.3226  & 6.5 & \boldred{27.07} & 0.3490\\
    \bottomrule
    \end{tabular}
    }
    \caption{Ablation for $\lambda_{rn}$, $\lambda_{fl}$ and $\lambda_{time}$. It is shown that our choice (in \textbf{bold}) leads to a better trade-off for both fidelity and realness. Best and second-best results shown in \boldred{red} and \boldblue{blue}.}
    \label{tab:ablation_l2}
    \vspace{-2mm}
\end{table}
\begin{table}
    \centering
    \resizebox{0.9\linewidth}{!}{
    \begin{tabular}{c|cccc}
    \toprule
        \textbf{Timestep} $t$ & \textbf{PSNR}$\uparrow$ & \textbf{LPIPS}$\downarrow$ & \textbf{NIQE}$\downarrow$ & \textbf{MUSIQ}$\uparrow$\\
        \hline
        0.0 & 24.34 & \boldred{0.3377} & \boldred{3.5455} & \boldred{69.52}\\
        0.2 & 24.45 & 0.3384 & 3.6803 & 69.25\\
        0.4 & 24.58 & 0.3397 & 3.8114 & 69.00\\
        0.6 & 24.72 & 0.3409 & 3.9368 & 68.60\\
        0.8 & 24.82 & 0.3423 & 4.0234 & 68.25\\
        1.0 & \boldred{24.85} & 0.3437 & 4.0438 & 67.96\\
    \bottomrule
    \end{tabular}
    }
    \caption{Results of the controllable trade-off between fidelity and realness, with adjustable properties implemented via timestep $t$. Test on the DIV2K validation set. Best results shown in \boldred{red}.}
    \label{tab:controllable}
    \vspace{-3mm}
\end{table}
\begin{table*}[]
\resizebox{\textwidth}{!}{%
\begin{tabular}{c|cccccccccc}
\toprule
\textbf{} & \textbf{StableSR}~\cite{StableSR_Wang_Yue_Zhou_Chan_Loy_2023} & \textbf{DiffBIR}~\cite{diffbir} & \textbf{SUPIR}~\cite{yu2024scaling} & \textbf{PASD}~\cite{yang2023pasd} & \textbf{ResShift}~\cite{yue2024resshift} & \textbf{InvSR}~\cite{yue2024arbitrary}  & \textbf{SinSR}~\cite{wang2024sinsr} & \textbf{OSEDiff}~\cite{wu2024osediff} & \textbf{Ours}  \\ \hline
\textbf{Inference Step} & 200 & 50 & 50 & 20 & 15 & \boldred{1}  & \boldred{1} & \boldred{1} & \boldred{1}  \\
\textbf{Inference Time} (s) & 12.4151 & 7.9637 & 16.8704 & 4.8441 & 0.7546 & \boldred{0.1416}  & \boldblue{0.1424} & 0.1791 & 0.1791  \\ 
\textbf{Total / Trainable Parameters} (M) & 1410 / 150.0 & 1717 / 380.0 & 2662.4 / 1331.2 & 1900 / 625.0 & \boldred{110} / 118.6 & 1010 / 33.8  &  \boldblue{119} / 118.6 & 1775 / \boldred{8.5} & 1775 / \boldred{8.5}  \\ 
\bottomrule
\end{tabular}%
}
\caption{Comparison of computational complexity, training time, and number of trainable parameters across diffusion-based methods. Loading time for model weights and dataloaders are not included. Best and second-best results are shown in \boldred{red} and \boldblue{blue}.}
\label{tab:time}
\vspace{-2mm}
\end{table*}
\begin{table}
    \centering
    \resizebox{0.9\linewidth}{!}{
    \begin{tabular}{c|cccc}
    \toprule
        \textbf{Method} & \textbf{PSNR}$\uparrow$ & \textbf{SSIM}$\uparrow$ & \textbf{LPIPS}$\downarrow$ & \textbf{Parameters} (M) $\downarrow$ \\
    \hline
        GSAD~\cite{hou2023global} & 28.67 & 0.9444 & 0.0487 & \boldred{17.17} \\
        Reti-Diff~\cite{he2023reti} & 27.53 & 0.9512 & 0.0349 & 26.11 \\
        GSAD (Distilled) & \boldred{28.69} & \boldred{0.9507} & \boldred{0.0336} & \boldred{17.17}\\
    \bottomrule        
    \end{tabular}
    \vspace{-4mm}
    }
\caption{Our distillation applied in low-light enhancement task evaluated on LOL-v2-syn~\cite{Chen2018Retinex} testset, which brings fidelity preservation and realness improvement. Best results shown in \boldred{red}.}
    \label{tab:lle}
    \vspace{-5mm}
\end{table}
\noindent \textbf{Quantitative Comparison}. We use RealESRGAN as a simulation of real-world degradation and compare the performance on the DIV2K, RealSR, and DRealSR validation sets. Tab.~\ref{tab:fidelity} and Tab.~\ref{tab:realness} present the quantitative comparison results.

Tab.~\ref{tab:fidelity} compares our method with existing methods that excel in terms of fidelity, showing that our method is comparable in terms of PSNR and SSIM, while significantly outperforming others on realness metrics such as DISTS, FID, and others. The comparison with RealESRGAN further demonstrates that diffusion-based methods generally achieve higher scores on no-reference metrics (NIQE, MANIQA, CLIPIQA, MUSIQ), suggesting that diffusion models are better suited to provide visual priors for super-resolution tasks.

Tab.~\ref{tab:realness} compares our method with existing methods that excel in realness. The results show that our method is competitive in realness metrics, while also achieving significant performance gains in fidelity.

\noindent \textbf{Qualitative Comparison}. Fig.~\ref{fig:comp} presents the results of comparison experiments on the RealSR testset. The figure shows that our method provides better visual quality and consistency with the original image compared to the other methods.

\noindent \textbf{Efficiency Comparison}.
To evaluate the efficiency and complexity of CTSR, we compare these properties with SOTA methods in Tab.~\ref{tab:time}, which shows that CTSR requires fewer inference steps, achieves comparable inference time and has fewer trainable parameters.


\subsection{Ablation Study}
\label{sec:ablation}

\noindent \textbf{Necessity of Teacher Distillation Loss}. A natural question arises: ``why do we need two teacher models to achieve the trade-off, given that many methods use $L_2$ loss and LPIPS loss for balancing fidelity and realness?". From a theoretical standpoint, $L_2$-norm, when used as the fidelity constraint, is too sparse and lacks the smoothness necessary to capture the detailed semantic information of the LR input. On the other hand, regularization losses like LPIPS struggle to effectively represent the distribution of natural images. By training SR models on a diffusion prior with various strategies, we can obtain better guidance for balancing fidelity and realness, thereby advancing the Pareto frontier of SR tasks. To further support this, we present results with and without the distillation loss in Tab.~\ref{tab:wot}. The comparison shows that without the distillation loss, the method reverts to the behavior of earlier GAN-based approaches, achieving better fidelity but suffering a significant decline in realness and visual quality.

\noindent \textbf{Selection of Teacher $\mathcal{T}_{fl}$}. Since multiple SOTA SR models excel in fidelity performance, to find the best choice for $\mathcal{T}_{fl}$, we also experiment with SinSR~\cite{wang2024sinsr} as the teacher model for dual teacher distillation. The results are presented in Tab.~\ref{tab:sinsr}.

\noindent \textbf{Selection of Coefficients $\lambda_{fl}$, $\lambda_{rn}$ and $\gamma_{time}$}. For the balancing coefficients among the loss function terms, we employ a grid search to determine the values that yield the best overall performance. The results of this selection process are shown in Tab.~\ref{tab:ablation_l2}.

\subsection{Evaluation of Controllability and Extendability} 
\noindent \textbf{Contollability}. Here, we introduce a controllable image super-resolution method enabled by the second stage of distillation that we propose. Specifically, the controllability of CTSR is determined by the input timestep $t$ of the diffusion model, where where $t=0$ corresponds to the best realness and $t=1$ corresponds to the best fidelity. The input $t$ can be selected anywhere between 0 and 1, allowing the user to adjust the balance between these two properties. We evaluate performance on the DIV2K validation set, with results presented in Tab.~\ref{tab:controllable}. As the input timestep $t$ increases from 0 to 1, fidelity metrics such as PSNR and SSIM improve, while realness metrics like LPIPS begin to decrease. Visual results are shown in Fig.~\ref{fig:teaser}(a) and \textbf{\textcolor{blue}{Supplementary Materials}}.

\noindent \textbf{Extension to Image Enhancement}. To demonstrate the generalization and versatility of our proposed fidelity-realness distillation method from Sec.
~\ref{sec:stage1}, we extend it to the low-light enhancement (LLE) task, showcasing the performance improvement achieved by this approach. We select two diffusion-based LLE methods: GSAD~\cite{hou2023global}, which excels in fidelity, and Reti-Diff~\cite{he2023reti}, which excels in realness, and apply a training strategy similar to our CTSR. The results, presented in Tab.~\ref{tab:lle}, show that our proposed distillation strategy preserves the fidelity advantage of GSAD while leveraging the model prior from Reti-Diff to enhance realness performance. To be specific, the performance improves by 0.02 in PSNR, 0.0063 in SSIM, and 0.0151 in LPIPS, thus validating the effectiveness and generality of our approach.

\section{Conclusion}
This paper proposes CTSR, a distillation-based real-world image super-resolution method that leverages multiple teacher models to strike a trade-off between realness and fidelity. Furthermore, inspired by the working pricinple of flow matching, to enable controllability between fidelity and realness, this paper explores a controllable trade-off effect by distilling the output distributions of the aforementioned models, enabling a controllable image super-resolution method that is able to be adjusted via input timestep. Experiments on several real-world image super-resolution benchmarks demonstrate the superior performance of CTSR, compared to other competing methods. Additionally, the proposed fidelity-realness distillation approach can be extended to other tasks like low-light enhancement for performance improvement.

\clearpage
\maketitlesupplementary

\noindent In the supplementary materials, we demonstrate additional experimental results, implementation details, discussion, and analysis as follows.


\section{More Implementation Details}
\label{sec: details}

\subsection{More Details of Loss Funtion}
\begin{figure*}
    \centering
    \includegraphics[width=0.6\linewidth]{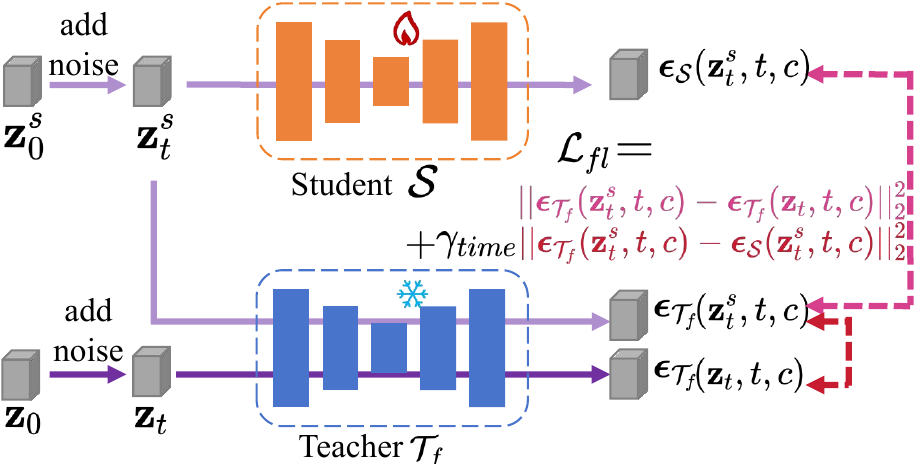}
    \caption{Visualized calculation process of $\mathcal{L}_{fl}$.}
    \label{fig:rn}
\end{figure*}

We provide a detailed loss calculation process for Stage 1 in the main paper, as shown in Fig.~\ref{fig:rn}.

\subsection{Pseudocode of Our Proposed CTSR Method}
The overall training process for first and second stage is shown in Algo.~\ref{alg:s1} and Algo.~\ref{alg:s2}.
    \begin{algorithm*}
    \caption{Fidelity-Realness Distillation in Stage 1}
    \label{alg:s1}
    \KwIn{Ground truth $\mathbf{x}_{GT}$, input LR image $\mathbf{x}_{LR}$, student model $\mathcal{S}$, teacher model $\mathcal{T}_{fl}$ and $\mathcal{T}_{rn}$, VAE encoder $\mathcal{E}$, VAE decoder $\mathcal{D}$, embedding of prompt $c$, loss balancing hyper-parameters $\lambda_{time}$, $\lambda_{fl}$, $\lambda_{rn}$, $\lambda_{l2}$, $\lambda_{lp}$}
    \KwOut{Student model $\mathcal{S}$}
    Initialize $\mathcal{S}$ using weight of $\mathcal{T}_{rn}$. \\
    \For{$epoch=1$ \KwTo $total \; epochs$}{
        $\mathbf{z}_{1} = \mathcal{E}(\mathbf{x}_{LR})$ \\
        $\mathbf{z}_{0} = \mathcal{E}(\mathbf{x}_{GT})$ \\
        Random sample a timestep $t$ \\
        $\mathbf{z}_{t} = add\_noise (\mathbf{z}_{0},t)$ \\
        $\mathbf{z}_{0}^s = \mathcal{S}(\mathbf{z}_{1})$ \\
        $\mathbf{x}_{0} = \mathcal{D}(\mathbf{z}_{0}^s)$ \\
        $\mathbf{z}_{t}^s = add\_noise (\mathbf{z}_{0}^s,t,c)$ \\
        $\mathcal{L}_{rec}=\lambda_{l2}||\mathbf{x}_{GT}-\mathbf{x}_{0}||_2^2+\lambda_{lp}\ell(\mathbf{x}_{GT},\mathbf{x}_{0})$ \\
        $\mathcal{L}_{fl}=||\boldsymbol{\epsilon}_{\mathcal{T}_{f}}(\mathbf{z}_t^s,t,c)-\boldsymbol{\epsilon}_{\mathcal{S}}(\mathbf{z}_t^s,t,c)||_2^2
        +\lambda_{time}||\boldsymbol{\epsilon}_{\mathcal{T}_{f}}(\mathbf{z}_t,t,c)-\boldsymbol{\epsilon}_{\mathcal{T}_{f}}(\mathbf{z}_t^s,t,c)||_2^2$ \\
        $\mathcal{L}_{rn}=||\boldsymbol{\epsilon}_{\mathcal{T}_{r}}(\mathbf{z}_t^s,t,c)-\boldsymbol{\epsilon}_{\mathcal{S}}(\mathbf{z}_t^s,t,c)||_2^2
        +\lambda_{time}||\boldsymbol{\epsilon}_{\mathcal{T}_{r}}(\mathbf{z}_t,t,c)-\boldsymbol{\epsilon}_{\mathcal{T}_{r}}(\mathbf{z}_t^s,t,c)||_2^2$ \\
        $\mathcal{L}_{s1}=\mathcal{L}_{rec}+\lambda_{fl}\mathcal{L}_{fl}+\lambda_{rn}\mathcal{L}_{rn}$  \\
        $\mathcal{L}_{s1}.backward()$ \\
        $\mathcal{S}.update()$ 
        
    }
    \Return $\mathbf{S}$
    \end{algorithm*}
  \begin{algorithm*}
    \caption{Controllability Distillation in Stage 2}
    \label{alg:s2}
    \KwIn{HR output of student model $\mathbf{x}_{0}$, student model $\mathcal{S}$, teacher model (weight initalized from student model) $\mathcal{T}_{\mathcal{S}}$, VAE encoder $\mathcal{E}$}
    \KwOut{Student model $\mathcal{S}$}
    \For{$epoch=1$ \KwTo $total \; epochs$}{
        Randomly sample timesteps $t$ and $t'$ $\in (0,1)$ /* ensure $t' > t$ */ \\
        $\mathbf{z}_{t}=\mathbf{z}_{0}+t\boldsymbol{\epsilon}_{\mathcal{T}_{\mathcal{S}}}(\mathbf{z}_0,t,c)$ \\
        $\mathbf{z}_{t'}=\mathbf{z}_{t}+t'\boldsymbol{\epsilon}_{\mathcal{T}_{\mathcal{S}}}(\mathbf{z}_0,t,c)$ \\
        
        $\mathcal{L}_{ctrl_{t,t'}}=||t\boldsymbol{\epsilon}_{\mathcal{T}_{\mathcal{S}}}(\mathbf{z}_t,t,c)-
        t'\boldsymbol{\epsilon}_{\mathcal{T}_{\mathcal{S}}}(\mathbf{z}_{t'},t',c)
        +(\Delta t )\boldsymbol{\epsilon}_\mathcal{S}(\mathbf{z}_t,t,c)||_2^2$ \\
        $\mathcal{L}_{s2}=\sum_{t,t' \in [0,1]}\mathcal{L}_{ctrl_{t,t'}}$ \\
        $\mathcal{L}_{s2}.backward()$\\
        $\mathcal{S}.update()$
    }
    \Return $\mathcal{S}$
    \end{algorithm*}


\section{More Experimental Results}
\label{sec: expr}

\subsection{More Results of Controllable Image SR}
Here we present the controllable image SR effect on the validation sets of DIV2K, RealSR and DRealSR. Results are shown in Tab.~\ref{tab:controllable_div2k}, Tab.~\ref{tab:controllable_realsr} and Tab.~\ref{tab:controllable_drealsr} seperately.
\begin{table*}
    \centering
    \resizebox{\linewidth}{!}{
    \begin{tabular}{c|ccccccccc}
    \toprule
        Timestep $t$ & PSNR$\uparrow$ & SSIM$\uparrow$ & LPIPS$\downarrow$ & DISTS$\downarrow$ & FID$\downarrow$ & NIQE$\downarrow$ & MUSIQ$\uparrow$ & MANIQA$\uparrow$ & CLIPIQA$\uparrow$\\
        \hline
        0.0 & 24.34 & 0.6093 & \boldred{0.3377} & \boldred{0.1377} & \boldred{24.56}  & \boldred{3.5455} & \boldred{69.52} & \boldred{0.5894} & \boldred{0.6741}\\
        0.2 & 24.45 & 0.6098 & 0.3384 & 0.1394 & 24.75  & 3.6803 & 69.25 & 0.5826 & 0.6726\\
        0.4 & 24.58 & 0.6131 & 0.3397 & 0.1412 & 25.00  & 3.8114 & 69.00 & 0.5767 & 0.6715\\
        0.6 & 24.72 & 0.6172 & 0.3409 & 0.1432 & 25.64  & 3.9368 & 68.60 & 0.5698 & 0.6684\\
        0.8 & 24.82 & 0.6191 & 0.3423 & 0.1447 & 26.13  & 4.0234 & 68.25 & 0.5642 & 0.6632\\
        1.0 & \boldred{24.85} & \boldred{0.6192} & 0.3437 & 0.1459 & 26.32  & 4.0438 & 67.96 & 0.5609 & 0.6585\\
    \bottomrule
    \end{tabular}
    }
    \caption{More results of the controllable trade-off between fidelity and realness, with adjustable properties implemented via timestep $t$. Test on the \textbf{DIV2K} validation set.}
    \label{tab:controllable_div2k}
\end{table*}

\begin{table*}
    \centering
    \resizebox{\linewidth}{!}{
    \begin{tabular}{c|ccccccccc}
    \toprule
        Timestep $t$ & PSNR$\uparrow$ & SSIM$\uparrow$ & LPIPS$\downarrow$ & DISTS$\downarrow$ & FID$\downarrow$ & NIQE$\downarrow$ & MUSIQ$\uparrow$ & MANIQA$\uparrow$ & CLIPIQA$\uparrow$\\
        \hline
        0.0 & 25.70 & 0.6962 & \boldred{0.3058} & \boldred{0.1530} & \boldred{121.30} & \boldred{4.0662} & \boldred{67.94} & \boldred{0.6367} & \boldred{0.6495}\\
        0.2 & 26.29 & 0.7211 & 0.3210 & 0.1620 & 127.67 & 4.2979 & 66.84 & 0.6314 & 0.6435\\ 
        0.4 & 26.61 & 0.7203 & 0.3178 & 0.1594 & 134.38  & 4.2320 & 66.33 & 0.6355 & 0.6340\\
        0.6 & 26.62 & 0.7204 & 0.3191 & 0.1605 & 145.21  & 4.2561 & 65.29 & 0.6340 & 0.6333\\
        0.8 & 26.65 & 0.7208 & 0.3206 & 0.1614 & 148.86  & 4.2708 & 62.64 & 0.6327 & 0.6240\\
        1.0 & \boldred{26.72} &  \boldred{0.7213} & 0.3220 & 0.1628 & 156.38  & 4.3209 & 61.08 & 0.6304 & 0.6209\\
    \bottomrule
    \end{tabular}
    }
    \caption{More results of the controllable trade-off between fidelity and realness, with adjustable properties implemented via timestep $t$. Test on the \textbf{RealSR} testset.}
    \label{tab:controllable_realsr}
\end{table*}

\begin{table*}
    \centering
    \resizebox{\linewidth}{!}{
    \begin{tabular}{c|ccccccccc}
    \toprule
        Timestep $t$ & PSNR$\uparrow$ & SSIM$\uparrow$ & LPIPS$\downarrow$ & DISTS$\downarrow$ & FID$\downarrow$ & NIQE$\downarrow$ & MUSIQ$\uparrow$ & MANIQA$\uparrow$ & CLIPIQA$\uparrow$\\
        \hline
        0.0 & 27.38 & 0.7767 & \boldred{0.3423} & \boldred{0.1937} & \boldred{142.52}  & \boldred{6.6438} & \boldred{64.70} & \boldred{0.6412} & \boldred{0.7060}\\
        0.2 & 27.53 & 0.7794 & 0.3446 & 0.1402 & 147.25 & 7.7594 & 63.52 & 0.6408 & 0.7042\\
        0.4 & 27.99 & 0.8023 & 0.3513 & 0.1687 & 150.39 & 7.5088 & 63.35 & 0.5654 & 0.6958\\
        0.6 & 28.22 & 0.8043 & 0.3528 & 0.2195 & 156.36 & 7.5306 & 62.99 & 0.5642 & 0.6930\\
        0.8 & 28.47 & 0.8056 & 0.3561 & 0.2369 & 161.24 & 7.8462  & 58.76 & 0.5453 & 0.6745\\
        1.0 & \boldred{28.68} & \boldred{0.8152} & 0.3697 & 0.2371 & 164.46 & 7.9699 & 57.85 & 0.5974 & 0.6664\\
    \bottomrule
    \end{tabular}
    }
    \caption{More results of the controllable trade-off between fidelity and realness, with adjustable properties implemented via timestep $t$. Test on the \textbf{DRealSR} testset.}
    \label{tab:controllable_drealsr}
\end{table*}

\subsection{More Visual Results}
\begin{figure*}
    \centering
    \includegraphics[width=1\linewidth]{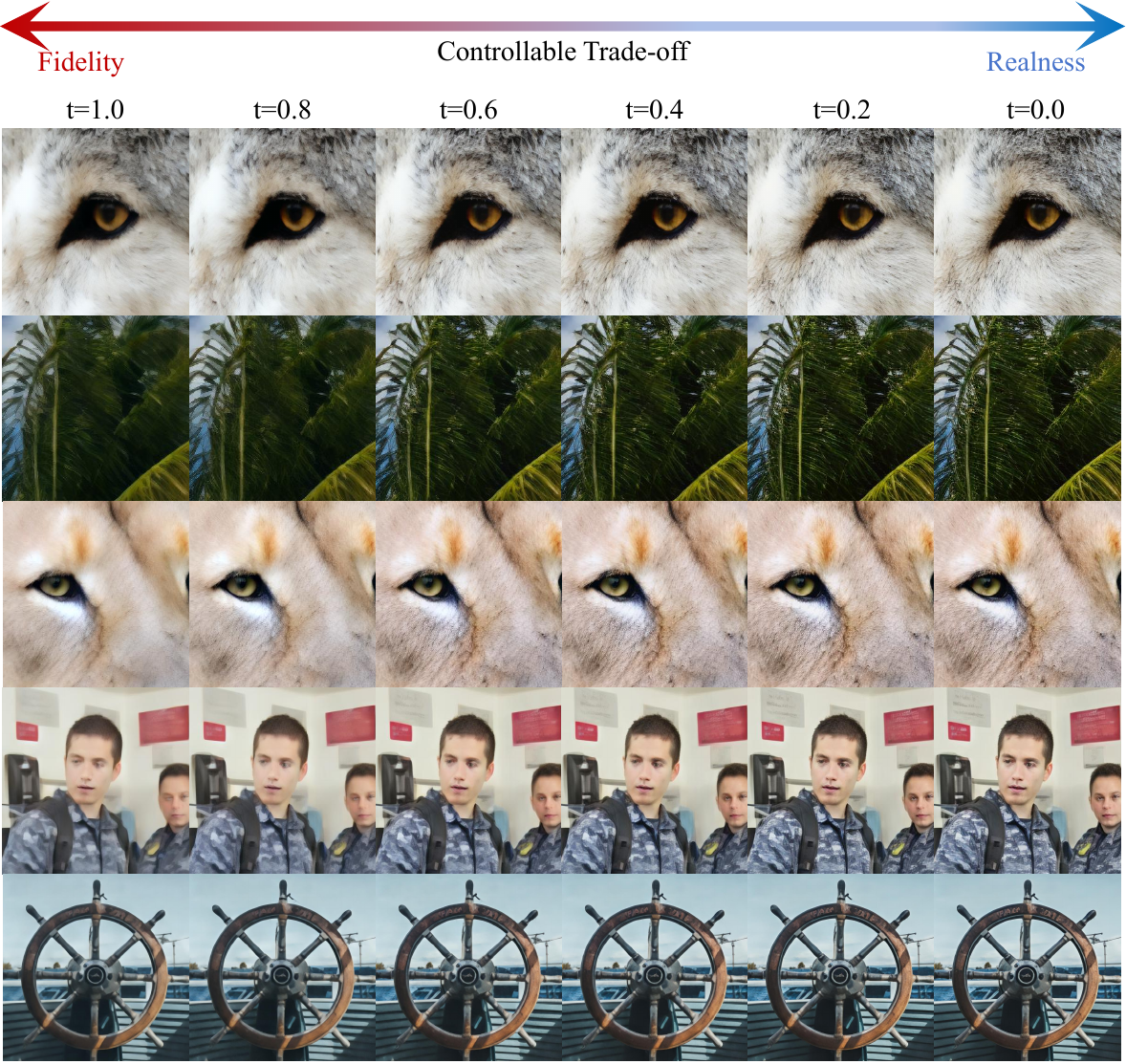}
    \caption{Visualized results of controllable image SR.}
    \label{fig:more}
\end{figure*}

We provide more results presenting the controllability of our proposed CTSR, which are shown in Fig.~\ref{fig:more}. From left to right, the fidelity property is gradually changed to realness, with less smooth and more details and better visual quality.

\clearpage

\small
\bibliographystyle{ieeenat_fullname}
\bibliography{main}


\end{document}